\documentclass[11pt,twocolumn]{article}

\usepackage[a4paper,margin=0.75in]{geometry}

\usepackage[T1]{fontenc}
\usepackage[utf8]{inputenc}
\usepackage{lmodern}
\usepackage{microtype}

\usepackage{amsmath, amssymb, amsthm, mathtools}

\usepackage{graphicx}
\usepackage{booktabs}
\usepackage{caption}
\usepackage{subcaption}
\usepackage{placeins}

\usepackage[numbers,sort&compress]{natbib}
\usepackage{xcolor}
\usepackage[
    colorlinks=true,
    linkcolor=blue,
    citecolor=blue,
    urlcolor=blue
]{hyperref}
\theoremstyle{definition}

\title{\Large\bfseries ChemQuests: A Curated Chemistry Question-Answer Database Extracted from ChemRxiv Papers}

\author{
    Mahmoud Amiri$^{1,2}$,
    Thomas Bocklitz$^{1,2}$
}

\date{\today}

\begin{document}

\twocolumn[
\begin{@twocolumnfalse}

\maketitle

\noindent
$^1$ Leibniz Institute of Photonic Technology, Member of Leibniz Health Technologies, Member of the Leibniz Centre for Photonics in Infection Research (LPI), Albert-Einstein-Strasse 9, 07745, Jena, Germany. \\
$^2$ Institute of Physical Chemistry (IPC) and Abbe Center of Photonics (ACP), Friedrich Schiller University Jena, Member of the Leibniz Centre for Photonics in Infection Research (LPI), Helmholtzweg 4, 07743, Jena, Germany. \\

\begin{abstract}

The rapid expansion of chemistry literature poses significant challenges for researchers seeking to efficiently access domain-specific knowledge. To support advancements in chemistry-focused natural language processing (NLP), we present ChemQuests, a curated dataset of 952 high-quality question-answer (QA) pairs derived from 155 ChemRxiv \cite{chemrxivWebsite} papers across 17 subfields of chemistry. Each QA pair is explicitly linked to its source text segment to ensure traceability and contextual accuracy. ChemQuests was constructed using an automated pipeline that combines optical character recognition (OCR), QA generation using GPT-4o, and fuzzy-search verification. The dataset emphasizes conceptual, mechanistic, applied, and synthetic or experimental questions, enabling applications in retrieval-based QA systems, search engine development, and fine-tuning of domain-adapted large language models. We analyze the dataset's structure, coverage, and limitations, and outline future directions for expansion and expert validation. ChemQuests provides a foundational resource for chemistry NLP research, education, and tool development.

\end{abstract}

\vspace{0.5em}
\noindent\textbf{Keywords:} Chemistry Question Answering, Natural Language Processing, Chemistry NLP Dataset, Large Language Models, Dataset Curation, Scientific Text Mining.

\vspace{1.5em}

\end{@twocolumnfalse}
]

\section{Introduction}

Chemistry is a diverse and rapidly evolving scientific discipline that is witnessing an unprecedented pace of research publications. Each year, millions of chemistry-related articles, reviews, and preprints are published across platforms such as peer-reviewed journals and open-access repositories like ChemRxiv \cite{lawlor2018preprints}. For example, ChemRxiv alone receives thousands of new submissions annually, reflecting the vibrant and fast-paced nature of chemical research. However, this growing body of literature presents significant challenges for chemists, researchers, and students seeking to efficiently locate, extract, and apply domain-specific knowledge.

Traditional literature search methods, which rely primarily on keyword-based queries through tools such as Web of Science \cite{webofscienceWebsite}, Google Scholar \cite{googlescholarWebsite}, and SciFinder \cite{scifinderWebsite}, often yield vast quantities of unstructured documents. These documents require labor-intensive manual review, lack semantic understanding, and rarely provide direct, contextualized answers to targeted chemical questions. Researchers are left to synthesize information themselves, limiting efficiency and accessibility, particularly for interdisciplinary or novice users.

Recent breakthroughs in NLP, especially large language models (LLMs) and retrieval-augmented generation (RAG), offer promising new approaches to address these challenges. LLMs such as GPT-3 and GPT-4 have demonstrated the ability to perform tasks ranging from text classification and question answering to autonomous experiment planning and synthesis optimization in chemistry and materials science contexts \cite{choi2024accelerating, boiko2023autonomous, jablonka2024leveraging, m2024augmenting}. Systems like Coscientist and ChemCrow exemplify how LLMs can be integrated with external tools and knowledge bases to automate experimental workflows, data extraction, and even discovery of novel compounds \cite{boiko2023autonomous, m2024augmenting}. These models significantly reduce the dependence on large annotated datasets or complex domain-specific architectures \cite{choi2024accelerating, jablonka2024leveraging}, lowering the barrier to entry for non-experts and accelerating knowledge-intensive research.
Despite these advances, chemistry-focused NLP continues to lag behind other fields, such as biomedicine \cite{miret2024llms}. This is due in part to the intrinsic complexity of chemical language, which features diverse terminologies, symbolic notations, and reaction mechanisms, and a lack of high-quality, structured, and curated datasets explicitly designed for training and evaluating chemistry-specific LLMs. Additionally, recent studies highlight the risks of hallucination and factual inaccuracies in generative models, which can undermine their utility in high-stakes scientific domains \cite{ji2023survey, farquhar2024detecting}. Approaches such as fine-tuning, entropy-based uncertainty estimation, and RAG have been proposed to mitigate these issues and improve answer reliability \cite{farquhar2024detecting, gao2023retrieval, zhang2024enhancing}.

To address these critical gaps, we introduce ChemQuests, a curated and systematically constructed QA dataset for chemistry-specific NLP applications. Built directly from primary literature on ChemRxiv, ChemQuests is designed to support robust question answering, improve retrieval precision, and serve as a fine-tuning or benchmarking resource for domain-specialized LLMs. Our aim is to catalyze advances in chemical information retrieval and automated understanding by providing a reliable, structured, and versatile dataset tailored to the unique demands of chemical language.

Research on scientific QA datasets and chemical NLP has laid an important foundation for the development of ChemQuests. Notable datasets in this space include PubMedQA \cite{jin2019pubmedqa}, BioASQ \cite{tsatsaronis2015overview}, SciQ \cite{welbl2017crowdsourcing}, and the Semantic Scholar Open Research Corpus (S2ORC) \cite{lo2019s2orc}. While these datasets have supported significant progress in biomedical and general scientific domains, they exhibit limitations when applied to the unique demands of chemistry-focused NLP.

PubMedQA focuses heavily on biomedical literature, offering QA pairs grounded in clinical contexts. Despite its strengths in medical applications, it does not capture the chemical specificity and breadth required for tasks in chemical informatics. Similarly, SciQ is constructed from educational assessment materials and is tailored toward general science education rather than research-grade chemical content, thus lacking the granularity needed for specialized QA systems. BioASQ, like PubMedQA, is centered on biomedical text mining and question answering, but its domain restriction excludes complex chemical representations. S2ORC, although extensive and multidisciplinary, remains loosely structured and lacks chemistry-targeted QA annotations, which hinders its utility for precise chemical information retrieval.

In the domain of chemical NLP, tools such as ChemDataExtractor \cite{swain2016chemdataextractor} and OSCAR \cite{jessop2011oscar4} have made significant advances in information extraction. ChemDataExtractor has been particularly effective in mining structured data such as molecular properties, synthesis conditions, and reaction outcomes, while OSCAR has improved chemical named entity recognition through rule-based and machine learning approaches. However, neither tool is designed to support structured QA tasks or generative modeling tailored to chemical discourse.

Despite the availability of these valuable resources and tools, a clear gap persists in the form of a curated, validated QA dataset that is specifically tailored to chemistry and directly derived from primary scientific literature. ChemQuests addresses this unmet need by providing an explicitly structured, chemistry-specific QA dataset with broad topical coverage and high-quality annotations suitable for downstream applications in chemical NLP.

In addressing the limitations outlined above, ChemQuests introduces a number of novel contributions to the field of chemical NLP. It represents the first large-scale, curated QA dataset specifically focused on chemical literature, sourced from ChemRxiv papers. The dataset comprises 952 QA pairs, each validated and explicitly linked to its original textual context to ensure traceability and scientific integrity.

The dataset construction process is underpinned by a rigorous and reproducible methodological pipeline. This includes OCR for accurate text extraction from Portable Document Format (PDF) documents, automated QA pair generation using the GPT-4o model, and a verification step employing fuzzy matching algorithms to ensure alignment between generated answers and their source contexts. This approach ensures that the resulting dataset is both accurate and reliable for use in high-stakes chemical NLP applications.

ChemQuests also provides comprehensive topical coverage across 17 subfields of chemistry, including organic, inorganic, physical, analytical, and materials chemistry. This diversity enhances its utility as a benchmark dataset and training resource for a wide array of tasks within the chemical sciences.
Beyond its content, ChemQuests is designed with practical applicability in mind. It can support the development of chemistry-specific retrieval-based QA systems, serve as a benchmark for evaluating domain-adapted NLP models, and aid in the fine-tuning of generative language models for scientific applications. Through these contributions, ChemQuests aims to significantly advance the capabilities of NLP systems in chemistry, enabling more effective access to and understanding of the chemical literature.
\section{Dataset Construction}

Creating ChemQuests involved a systematic and clearly structured pipeline comprising several integral stages: data source selection, comprehensive text extraction and preprocessing, generation of high-quality QA pairs using advanced language models, and rigorous answer verification and indexing. This section outlines each of these stages to ensure methodological transparency and reproducibility.

\subsection{Data Source: ChemRxiv papers}

\begin{figure*}[t]
    \centering
    \includegraphics[width=1\textwidth]{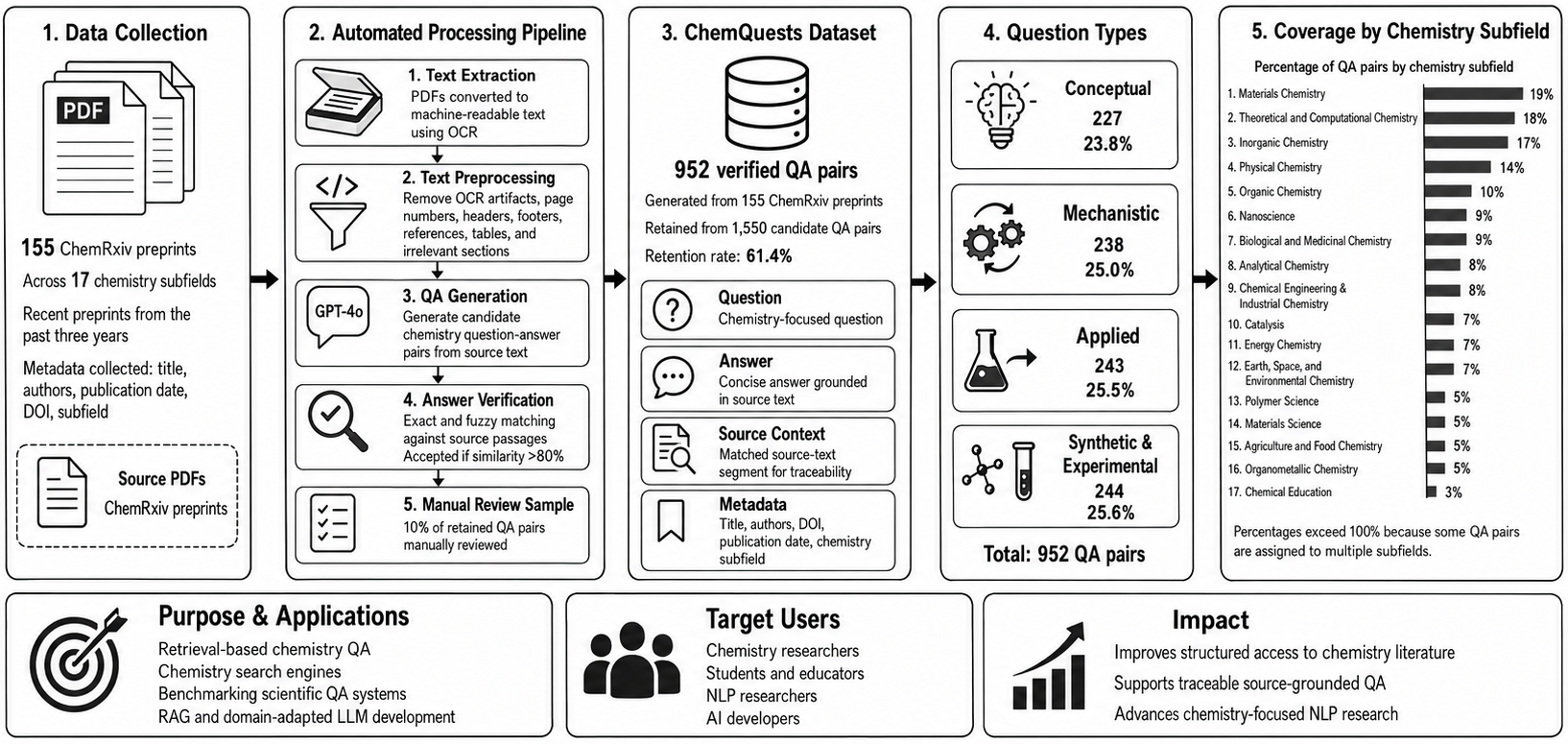}
    \caption{Infographic summarizing the ChemQuests dataset construction workflow and dataset characteristics. The figure illustrates the collection of 155 ChemRxiv papers across 17 chemistry subfields, automated OCR-based text extraction, GPT-4o-assisted question-answer generation, fuzzy-search verification, and the resulting curated dataset of 952 validated chemistry QA pairs. It also summarizes the distribution of question types, chemistry subfield coverage, and key applications, including retrieval-based QA systems, chemistry search engines, and fine-tuning of domain-adapted large language models.
}
    \label{fig:ChemQuests_pipeline}
\end{figure*}
We selected ChemRxiv, a prominent open-access repository for chemistry research, as the sole data source for ChemQuests. ChemRxiv provides a broad collection of recent, high-quality papers across many subfields of chemistry and operates under the Creative Commons CC BY \cite{creativecommonsCCBY40} license, which explicitly permits unrestricted reuse and adaptation, making it well suited for constructing an open dataset.

To ensure disciplinary diversity and thematic balance, we curated a total of 155 papers, evenly distributed across 17 well-established subfields of chemistry: Agriculture and Food Chemistry, Analytical Chemistry, Biological and Medicinal Chemistry, Catalysis, Chemical Education, Chemical Engineering and Industrial Chemistry, Earth, Space, and Environmental Chemistry, Energy Chemistry, Inorganic Chemistry, Materials Chemistry, Materials Science, Nanoscience, Organic Chemistry, Organometallic Chemistry, Physical Chemistry, Polymer Science, and Theoretical and Computational Chemistry.

papers were selected based on three criteria: recency (published within the past three years), subfield balance (ensuring roughly equal representation from each category), and quality (emphasizing clarity of presentation, depth of chemical content, and potential for generating meaningful conceptual or mechanistic questions). Metadata such as publication date, title, authors, and fields of study were automatically extracted using OpenAI's GPT-3.5-turbo model and validated against a controlled vocabulary. This curated collection ensures a dataset that reflects both the breadth and depth of contemporary chemical research.

\subsection{Text Extraction and Preprocessing}

As most ChemRxiv papers are distributed as PDFs, accurate text extraction was essential. To achieve this, we used olmOCR \cite{poznanski2025olmocr}, an open-source OCR tool specifically optimized for scientific literature. This tool leverages convolutional neural networks (CNNs) trained on scientific texts, ensuring high fidelity in recognizing technical terminology and chemical nomenclature.

Each PDF was processed individually at 300 dots per inch (DPI) resolution, converting scanned pages into machine-readable text. The output was then cleaned using Python scripts to remove OCR artifacts (e.g., misrecognized characters and redundant whitespace), along with extraneous content such as page numbers, headers, and footers.

Subsequently, we filtered out irrelevant content such as figure captions, tables, and reference sections. This was achieved using regex-based heuristics tailored to common formatting patterns in scientific writing. To ensure accuracy, a randomly selected 10\% sample of documents was manually reviewed and used to iteratively refine the heuristics. The resulting preprocessed text was of sufficient quality for reliable QA generation.

\subsection{Question-Answer Generation Process}

QA pairs were generated using OpenAI's GPT-4o model due to its strong performance on scientific reasoning tasks. The model was prompted with structured instructions to generate three specific types of questions: (1) conceptual questions explaining fundamental chemistry concepts, (2) mechanistic questions describing chemical processes, and (3) applied questions discussing the practical implications of the research. The retained QA pairs were later classified into four evaluation categories, including synthetic and experimental questions.

Each prompt required the model to return a clearly stated question, a concise answer, and a directly relevant textual snippet quoted verbatim or nearly verbatim from the source papers. This structure was designed to ensure that every QA pair was accurate, grounded in source material, and easy to validate.

To optimize output reliability, the GPT-4o model was configured with a temperature of 0.2 to minimize randomness. For each of the 155 papers, 10 QA pairs were generated, yielding a total of 1,550 candidate pairs.

\subsection{Answer Verification and Indexing}

To ensure factual alignment and traceability, each generated answer was validated by checking its similarity to the original source text. A hybrid verification pipeline was implemented:

First, an exact substring match was attempted. If that failed, we applied approximate matching using the fuzzysearch and rapidfuzz libraries, leveraging Levenshtein-based edit distance \cite{navarro2001guided} and partial ratio scores \cite{rao2018partial}. A match was accepted if its similarity exceeded an empirical threshold of 80\%.

A sliding window approach (400-character windows, 100-character steps) was used to localize the best-matching passage. Matched passages were expanded by 10 characters on each side to capture partial truncations. QA pairs failing to meet the threshold were discarded.

A random 10\% of the retained QA pairs were manually reviewed. This manual validation confirmed an estimated post-filtering accuracy exceeding 95\%. From the original pool of 1,550 generated QA pairs, 952 passed the verification pipeline, yielding a 61.4\% retention rate.
\section{Dataset Analysis}
This section presents a comprehensive quantitative and qualitative analysis of the ChemQuests dataset. The dataset comprises 952 validated QA pairs, extracted from 155 ChemRxiv papers. We analyze the distribution across chemical subfields and categorize the question types to demonstrate the dataset's diversity and utility.
\subsection{Distribution Across Chemistry Subfields}
To ensure broad representativeness, the dataset was stratified to provide substantial coverage across 17 major subfields of chemistry. As shown in Table~\ref{tab:subfields}, the distribution reflects the interdisciplinary nature of chemical research, where some QA pairs belong to multiple subfields.
\begin{table*}[t]
    \centering
    \scriptsize
    \caption{Percentage of QA pairs per chemistry subfield in ChemQuests}
    \label{tab:subfields}
    \begin{tabular}{l c}
    \toprule
    \textbf{Chemistry Subfield} & \textbf{Percentage (\%)} \\
    \midrule
    Materials Chemistry & 19 \\
    Theoretical and Computational Chemistry & 18 \\
    Inorganic Chemistry & 17 \\
    Physical Chemistry & 14 \\
    Organic Chemistry & 10 \\
    Nanoscience & 9 \\
    Biological and Medicinal Chemistry & 9 \\
    Analytical Chemistry & 8 \\
    Chemical Engineering \& Industrial Chemistry & 8 \\
    Catalysis & 7 \\
    Energy Chemistry & 7 \\
    Earth, Space, and Environmental Chemistry & 7 \\
    Polymer Science & 5 \\
    Materials Science & 5 \\
    Agriculture and Food Chemistry & 5 \\
    Organometallic Chemistry & 5 \\
    Chemical Education & 3 \\
    \bottomrule
    \end{tabular}
    \caption*{\scriptsize Percentages exceed 100\% because some QA pairs are assigned to multiple subfields.}
\end{table*}

On average, each subfield contains approximately 87 QA pairs, with a standard deviation of around 47.6, indicating moderate variation in distribution. This stratification ensures that the dataset encompasses a wide range of chemical domains, from theoretical principles to applied materials science.
\subsection{Analysis of Question Types}

To investigate the diversity of the content more thoroughly, each QA pair in the dataset was classified into one of four primary categories, based on the conceptual scope and focus of the question. These categories include conceptual, mechanistic, applied, and synthetic or experimental questions. Conceptual questions are those that address fundamental principles or theoretical definitions, such as inquiries into the nature of chemical interactions-for example, ``What defines a Lewis acid-base interaction?'' Mechanistic questions, on the other hand, delve into the specifics of chemical processes or reaction pathways, exemplified by questions like ``How does a palladium catalyst facilitate cross-coupling reactions?'' Applied questions emphasize real-world implementations and technological relevance, as seen in examples such as ``What are potential applications of graphene oxide in energy storage?'' Finally, synthetic and experimental questions focus on laboratory methods, reagents, and protocols, such as ``Which reagent is commonly used for selective oxidation of alcohols?''

The distribution of these question types is summarized in Table~\ref{tab:qtypes}. Synthetic and experimental questions constitute the largest portion of the dataset, accounting for 25.6\% of all QA pairs, and provide essential support for real-world chemical workflows and laboratory-oriented retrieval systems. Applied questions represent 25.5\% of the dataset, followed by mechanistic questions at 25.0\% and conceptual questions at 23.8\%, illustrating the dataset's alignment with both theoretical insight and practical utility.

\begin{table*}[t]
\centering
\scriptsize
\caption{Distribution of QA Pair Types in ChemQuests}
\label{tab:qtypes}
\begin{tabular}{lcc}
\toprule
\textbf{Question Type} & \textbf{Number of Questions} & \textbf{Percentage (\%)} \\
\midrule
Conceptual & 227 & 23.8 \\
Mechanistic & 238 & 25.0 \\
Applied & 243 & 25.5 \\
Synthetic and Experimental & 244 & 25.6 \\
\midrule
\textbf{Total} & 952 & 100.0 \\
\bottomrule
\end{tabular}
\end{table*}

Although the source were selected to balance subfields, some domains naturally yield more accessible or conceptually focused content, which may influence the type and complexity of the generated QA pairs. For instance, conceptual questions are more readily derived from educational or review-type texts, whereas experimental fields may pose challenges in question generation due to technical density.
Future expansions of ChemQuests may include expert-driven annotations to further assess and mitigate these biases. Incorporating difficulty ratings or ambiguity scores could enhance the dataset's granularity and support more nuanced model evaluations.

\section{Applications and Use Cases}

ChemQuests is designed as a versatile and robust resource for advancing a wide range of chemistry-specific NLP applications. Its structured and validated QA format, along with explicit linkage to source papers, enables numerous practical uses across research, development, education, and benchmarking. This section outlines key application areas where ChemQuests provides direct value.

\subsection{Retrieval-Based Chemistry Search Engines}

ChemQuests can serve as a valuable testbed for developing and evaluating chemistry-specific retrieval-based search systems. Given its grounding in real scientific literature, the dataset supports the development of domain-aware information retrieval frameworks that go beyond keyword-based matching. By leveraging the explicit question-context-answer structure, researchers can benchmark retrieval models (e.g., Best Match 25 (BM25) and dense vector search with Facebook AI Similarity Search (FAISS)) in terms of recall@k, mean reciprocal rank (MRR), and contextual relevance. Furthermore, its integration with semantic retrieval systems using LLMs or embedding-based methods enables evaluation of hybrid retrieval-generation pipelines in real-world chemistry scenarios.

\subsection{Fine-Tuning Chemistry-Specific Large Language Models}

Domain-specific fine-tuning is essential for improving the performance of large language models on specialized tasks where general pretraining falls short. In this context, ChemQuests offers a high-quality corpus for supervised fine-tuning of chemistry-focused LLMs. Researchers aiming to create customized models, such as a domain-adapted variant of GPT (e.g., ``ChemGPT''), can utilize the dataset's rich and diverse QA pairs to instill deeper domain knowledge into the model.

The fine-tuning process involves using chemistry-specific questions as input prompts and their corresponding validated answers as outputs. For instance, a prompt like ``Explain the role of Lewis acids in Friedel-Crafts reactions'' would be paired with an accurate, contextually relevant answer sourced from a ChemRxiv papers. Training can be implemented through frameworks such as Hugging Face Transformers or the OpenAI fine-tuning API. Performance metrics such as Bilingual Evaluation Understudy (BLEU), Recall-Oriented Understudy for Gisting Evaluation (ROUGE), and exact match accuracy offer quantitative evaluation, while the explicit association of each QA pair with source snippets enables integration with advanced techniques like RAG and reinforcement learning from human feedback (RLHF).

\subsection{Benchmarking NLP Models in Scientific Question Answering}

Beyond its role as a training corpus, ChemQuests functions as a rigorous benchmark for scientific question answering in the chemistry domain. Its validated and source-grounded QA pairs make it ideal for evaluating NLP models under standardized and reproducible conditions. Researchers can assess a variety of model architectures, including encoder-decoder transformers, retrieval-only models, and RAG-based hybrids, by measuring their performance on chemistry-specific tasks.

Standard evaluation procedures include dataset splits into training, validation, and test sets, with metrics such as exact match (EM), ROUGE-L, and BLEU providing insight into both lexical and semantic accuracy. The domain richness of the dataset further allows for fine-grained evaluation by question type, subfield (e.g., organic, inorganic, materials chemistry), or reasoning complexity. These capabilities support the systematic analysis of model strengths and weaknesses, informing the development of more robust and chemically literate NLP systems.

\subsection{Evaluating Embedding Models for Scientific Text Retrieval}

The dataset also enables systematic benchmarking of text embedding models in scientific and chemical contexts. Because each question in ChemQuests is associated with a specific answer and its source document, researchers can assess how well various embedding models perform in retrieving semantically relevant contexts. Models such as SciBERT, SPECTER, E5, and OpenAI or Cohere embeddings can be evaluated using metrics like recall@k, normalized discounted cumulative gain (nDCG), and semantic similarity to the gold answer.

This application is particularly valuable in designing high-performance retrieval layers in RAG systems or chemistry-aware semantic search tools. By comparing embedding models across a common evaluation framework, the dataset facilitates a clearer understanding of which vector representations are best suited for chemical language.

\subsection{Educational Applications in Chemistry}

ChemQuests also holds significant value for chemical education. Its structured QA format supports a variety of pedagogical tools, from interactive learning platforms and automated tutoring systems to quiz generation and self-assessment modules. Because each QA pair is validated and linked to a credible scientific source, it offers students and educators trustworthy material for exploring complex concepts, mechanisms, and applications in modern chemistry. Interactive systems can use these QA pairs to simulate question-driven inquiry or Socratic learning environments, encouraging deeper conceptual engagement.

\subsection{Analyzing Knowledge Drift and Domain Adaptation in LLMs}

Finally, ChemQuests enables longitudinal studies on how large language models evolve over time in their knowledge of chemistry. By re-evaluating model outputs across different LLM versions (e.g., GPT-3.5 vs. GPT-4 vs. domain-fine-tuned variants), researchers can track changes in factual accuracy, terminology usage, and reasoning quality. This is particularly relevant in assessing knowledge drift, where model responses may shift as a function of pretraining data updates or architectural changes.

Additionally, the dataset provides a platform for measuring the impact of domain adaptation efforts. Models pre-trained on general corpora can be compared to models fine-tuned on ChemQuests to determine improvements in answer correctness, retrieval fidelity, and reasoning coherence in chemical domains.

\section{Limitations}
Despite its scope and utility, the ChemQuests dataset exhibits several notable limitations. A primary concern stems from the reliance on LLMs to automatically generate questions and answers. Although the generation process was carefully designed to minimize ambiguity and speculative reasoning, instances of model hallucination remain a challenge. In certain cases, LLM-generated answers referenced chemical reactions or mechanistic details not explicitly supported by the original paper content, introducing plausible-sounding but incorrect information. These inaccuracies, if unchecked, could propagate misinformation and compromise the dataset's scientific reliability. Moreover, the current answer verification approach, which is based on fuzzy token matching via Levenshtein distance with a fixed similarity threshold, prioritizes textual similarity over semantic correctness. As a result, syntactically accurate but contextually invalid answers may inadvertently be accepted into the final dataset.

Another significant limitation concerns inherent biases and imbalances within the dataset. Although ChemQuests spans 17 chemistry subfields, the distribution is not uniform. Subfields such as Materials Chemistry, Theoretical and Computational Chemistry, and Inorganic Chemistry are disproportionately represented. In contrast, areas like Agriculture and Food Chemistry and Chemical Education remain underrepresented. This imbalance could introduce bias during model training, resulting in uneven performance and limiting the generalizability of the dataset across the broader spectrum of chemical disciplines.

A further limitation arises from the exclusive use of ChemRxiv papers as source material. While papers facilitate rapid and open dissemination of emerging research, they have not undergone formal peer review, raising questions about the reliability, completeness, and scientific rigor of the information they contain. Claims or mechanistic insights presented in these papers may be revised, corrected, or even retracted upon peer review. Consequently, the reliance on paper literature introduces an additional layer of uncertainty regarding the long-term validity and accuracy of QA pairs derived from such sources.

\section{Future Work}
To address current limitations and further enhance the quality and utility of the ChemQuests dataset, several avenues for future development are envisioned. A central priority involves incorporating human expert validation into the dataset construction process. By involving domain experts from various subfields of chemistry, a portion of the QA pairs can undergo rigorous manual review, allowing for in-depth accuracy assessment. Such validation would not only help quantify the frequency and nature of LLM hallucinations but also provide critical feedback for refining low-quality content. Integrating human-in-the-loop mechanisms, where expert feedback is directly used to revise or filter questions and answers, could markedly improve the reliability and trustworthiness of the dataset.

Another key direction is addressing content imbalances across different chemistry subfields. Currently underrepresented areas such as Agriculture and Food Chemistry, Environmental Chemistry, and Chemical Education may benefit from targeted curation strategies. Employing stratified sampling approaches based on subfield-specific quotas could help ensure that future dataset releases offer a more comprehensive and balanced representation of chemical knowledge. This intentional diversification would make ChemQuests more inclusive and valuable for a wider range of applications.

In addition, expanding the dataset to include content from peer-reviewed literature represents a significant opportunity to enhance quality and credibility. Future work will focus on integrating material from reputable open-access sources such as PubChem, Europe PMC, and major chemistry journals. This expansion will involve developing criteria for selecting high-quality, recent publications from diverse subfields, thereby ensuring that the dataset reflects both academic rigor and topical relevance.

Improving the semantic verification of QA pairs is also a priority. Current methods, such as fuzzy string matching, may fail to capture nuanced semantic inaccuracies. Future iterations of ChemQuests will explore the application of more advanced semantic matching techniques, including transformer-based models and sentence-level embeddings like Sentence-BERT. These approaches are expected to increase the precision of answer validation, thereby reducing the inclusion of subtly incorrect or misleading QA pairs. Comparative benchmarking of different semantic verification strategies will guide the selection of the most robust method for integration.

Finally, the sustainability and long-term impact of ChemQuests will be supported through continuous updates and active community involvement. Establishing collaborative platforms, such as GitHub repositories and annotation tools, will enable open contributions, transparent validation workflows, and regular version-controlled releases. This participatory approach will ensure that ChemQuests remains a dynamic, community-driven resource, adaptable to evolving research needs in both chemistry and natural language processing.

\section{Conclusion}

The paper introduces \textbf{ChemQuests}, a structured and validated QA dataset designed to support NLP in chemistry. It was developed by extracting and processing content from 155 ChemRxiv papers across 17 chemistry subfields, resulting in 952 carefully verified QA pairs. ChemQuests addresses the lack of chemistry-specific NLP resources and serves as a valuable asset for improving retrieval-augmented systems, fine-tuning large language models, and evaluating scientific knowledge extraction techniques.

The dataset was built using an automated pipeline that integrates OCR, GPT-based QA generation, and fuzzy-search verification. This pipeline, which is detailed and validated in the paper, can be adapted for constructing domain-specific QA datasets in other scientific fields such as physics or engineering. We emphasize the dataset's quality, achieved through a similarity-based validation approach that supports relevance and accuracy, both of which are critical for trustworthy NLP applications.

Despite its strengths, ChemQuests has notable limitations. We acknowledge the possibility of hallucinated answers that do not precisely align with source texts, which may require manual review for high-stakes use. 

To address these issues, future plans for ChemQuests include expanding the dataset using peer-reviewed literature, incorporating human expert validation to further ensure accuracy, and regularly updating the dataset to keep pace with new developments in the field. We envision a dynamic, community-driven resource that evolves over time.

Overall, ChemQuests represents a useful contribution to the field of chemistry-focused artificial intelligence (AI), offering researchers, educators, and industry professionals a reliable and structured source of chemical information. It paves the way for more context-aware NLP systems and sets a benchmark for scientific QA datasets in chemistry.

\section{Data Availability}

The ChemQuests dataset is publicly available on Hugging Face at \url{https://huggingface.co/datasets/Bocklitz-Lab/ChemQuests}.

\section{Competing Interests}

The authors declare no competing interests.

\section{Funding}

This work is supported by the BMFTR, funding program Photonics Research Germany (13N15466 (LPI-BT1-FSU), 13N15710 (LPI-BT3-FSU), 13N15715 (LPI-BT4-FSU)) and is integrated into the Leibniz Centre for Photonics in Infection Research (LPI). The LPI, initiated by Leibniz IPHT, Leibniz-HKI, Friedrich Schiller University Jena, and Jena University Hospital, is part of the BMFTR national roadmap for research infrastructures.

\begingroup
\footnotesize
\bibliographystyle{plainnat}
\bibliography{ref}
\endgroup

\end{document}